\documentclass[runningheads,a4paper]{llncs}

\usepackage{amssymb}
\usepackage{graphicx}

\usepackage{url}
\urldef{\mailsa}\path|{vivian.santossilva, andre.freitas, siegfried.hasdschuh}@uni-passau.de|   
\newcommand{\keywords}[1]{\par\addvspace\baselineskip
\noindent\keywordname\enspace\ignorespaces#1}

\begin{document}

\title{Word Tagging with Foundational \\Ontology Classes: Extending the \\WordNet-DOLCE Mapping to Verbs}

\titlerunning{Word Tagging with Foundational Ontology Classes}

\author{Vivian S. Silva
\thanks{CNPq Fellow -- Brazil.}
\and Andr\'{e} Freitas\and Siegfried Handschuh}

\institute{Department of Computer Science and Mathematics \\ University of Passau \\ Innstra\ss{}e 43, 94032, Passau, Germany \\
\mailsa\\}

\maketitle

\begin{abstract}
Semantic annotation is fundamental to deal with large-scale lexical information, mapping the information to an enumerable set of categories over which rules and algorithms can be applied, and foundational ontology classes can be used as a formal set of categories for such tasks. A previous alignment between WordNet noun synsets and DOLCE provided a starting point for ontology-based annotation, but in NLP tasks verbs are also of substantial importance. This work presents an extension to the WordNet-DOLCE noun mapping, aligning verbs according to their links to nouns denoting perdurants, transferring to the verb the DOLCE class assigned to the noun that best represents that verb's occurrence. To evaluate the usefulness of this resource, we implemented a foundational ontology-based semantic annotation framework, that assigns a high-level foundational category to each word or phrase in a text, and compared it to a similar annotation tool, obtaining an increase of 9.05\% in accuracy.
\end{abstract}

\section{Introduction}\label{intro}
Lexical semantic information is fundamental in many natural language processing and semantic computing applications \cite{ciaramita2003supersense}, \cite{pasca2001high}, \cite{pustejovsky2002medstract}. Applications such as question answering and text entailment require complex inferences involving large commonsense knowledge bases. The consumption, interpretation and coordination over large-scale lexical information demands the use of higher level categories capable of generalizing the information without loss in meaning.

The large symbolic word space which is the target of NLP tasks demands strategies to map words to higher level classes which are enumerable and can be used to encode rules and algorithms on the top of these classes. The utility of tagging lies on the potential for encoding generalizations using an enumerable set of categories. These categories can range from simple lexical information, like the grammatical class of a word, to more complex semantic representation, intended to unambiguously state what a concept means in the world. Foundational ontology classes are a good example of semantic representation, composing a set of categories that can determine the most high-level nature of a concept. Additionally, the foundational ontology entities and their connection to logics supports the connection between natural language and reasoning. This connection can help NLP systems to address complex semantic interpretation tasks.

WordNet \cite{fellbaum1998wordnet} can be used as a ``bridge'' between natural language text and higher level semantic representations, including the foundational ontology-based modelling. In an effort to provide the WordNet taxonomy with more rigorous semantics, Gangemi et al. \cite{gangemi2003sweetening} performed the alignment between WordNet upper level synsets and the foundational ontology DOLCE \cite{masolo2003wonderweb}. After a meticulous analysis, the WordNet taxonomy was reorganized to meet the OntoClean \cite{guarino2009overview} methodology requirements, and the resulting upper level nouns were then mapped to DOLCE classes representing their highest level categories. This mapping concentrated on the noun database, since most particulars in DOLCE describe categories whose members are denoted by nouns.

On the other hand, many NLP applications need to deal with events, actions, states, processes and other temporal entities that may not be expressed as nouns, but rather as verbs. Often, verbs are seen as relationships between concepts, and DOLCE in fact provides a well-defined set of properties and axioms that link classes together in a meaningful way, like the properties \textit{performs}, \textit{target}, \textit{instrument}, \textit{makes} or \textit{uses}, among many others. But in natural language a verb can play the role of an entity itself, and a class will be more suitable to represent it than a property. 

As an example, consider a rule-based text entailment task where, given the fact ``Mary is a mother'', known to be true, we want to check whether the fact ``Mary gave birth'' is also true. Mapping the terms to foundational ontology classes can be done as an intermediary step to reduce the reasoning search space. Here, ``give birth'' would be better classified as an \textit{action}, while ``Mary'' can be seen as an \textit{agent}, and ``mother'' as a \textit{role}. Using a supporting definition (for example, from WordNet) stating that ``a mother is a woman who has given birth'' and a pre-defined rule asserting that ``\textit{if} an agent plays a role \textit{and} the role performs an action, \textit{then} the agent performs the action (while playing the role)'', we would have that ``\textit{if} Mary plays the role of a mother, \textit{and} a mother performs the action of giving birth, \textit{then} Mary gave birth''. As can be seen, the classification of the verb ``give birth'' as a member of a foundational category is crucial for applying the correct rule and accomplishing the entailment.

This work aims at proposing a semantic annotation model based on foundational ontologies, called FO Tagging, that can be used to enrich text from a knowledge base, bringing valuable information to the execution of natural language processing tasks and semantic computing, and reducing the size of symbolic space they need to deal with. To accomplish this goal, we present an alignment between WordNet verbs and DOLCE, taking as starting point the nouns alignment provided by Gangemi et al. \cite{gangemi2003sweetening}. To identify the correct class for each top level verb, we start by searching for direct links between those verbs and their noun counterparts, that is, a noun that represents an occurrence, or \textit{eventuality} in the Davidsonian logical view \cite{davidson1967logical}, of that verb. When there are no such direct links, either a path between the verb and a suitable noun is drawn through indirect links, or a manual evaluation based on the terms present in the synset's gloss is carried out. The DOLCE classes assigned to the top level verbs are then propagated down the taxonomy, resulting in a fully classified verb database.

The paper is organized as follows: Sec. \ref{relwork} lists some related work regarding the link between linguistic resources and foundational ontologies, as well as semantic annotation approaches. Section \ref{verbonto} presents the basis for the ontological structuring of verbs. Section \ref{method} details the methodology adopted in the alignment followed by the results in Sec. \ref{alignres} and a quantitative evaluation in Sec. \ref{eval}. Section \ref{concl} draws the conclusions and points to future work.

\section{Related Work}\label{relwork}
The alignment between WordNet noun synsets and the foundational ontology DOLCE performed by Gangemi et al. \cite{gangemi2003sweetening} is probably the most comprehensive attempt to turn WordNet into a conceptually well-grounded ontology. Since DOLCE was developed under a rigorous methodology that ensures the consistency across the taxonomy links, and is oriented towards human cognition and natural language, the resulting mappings and reorganized noun hierarchy can potentially be more useful in practical applications.

To carry out the alignment, the noun synsets taxonomy was analyzed taking into account a set of criteria such as identity, rigidity and unity \cite{guarino2009overview}, as well as concepts and individuals differentiation, generality level, among others. The identified inconsistencies were corrected with synsets exclusion or relocation, and the selected top synsets were then mapped to DOLCE classes, adding an upper level descriptive layer to the WordNet ontology.

Another effort to map WordNet concepts to a foundational ontology was presented by Niles and Pease \cite{niles2003linking}, who manually mapped all the WordNet synsets to the Standard Upper Merged Ontology \cite{pease2002suggested}. SUMO can be better defined as a knowledge base rather than a pure foundational ontology, because, differently from DOLCE, it contains many domain specific concepts besides the upper level, domain independent classes. This characteristic leads to a different kind of alignment, where the primary goal was to link the synsets to a similar class in the ontology, that is, a class that has the same meaning, and not a class that represents their upper level category. Only when a similar class was not found in the ontology, a class showing other kind of relationship, like subsumption or instantiation, was chosen and assigned to the synset.

Although all the verb synsets were also mapped to SUMO, the final classification is very heterogeneous. For example\footnote{Mappings available at https://goo.gl/bflXqx}, one of the senses of the verb \textit{breathe} (``draw air into, and expel out of, the lungs'') was mapped to the SUMO concept \textit{Breathing}, considered equivalent in meaning, but the verb \textit{palpebrate} (``wink or blink, especially repeatedly'') was mapped to the higher level concept \textit{PhysiologicProcess}, which subsumes it. Even concepts that don't represent temporal entities were assigned to verbs, like the first sense of the verb \textit{sigh} (``heave or utter a sigh; breathe deeply and heavily''), which was mapped to the concept \textit{Organism}, considered equivalent in meaning. The work presented in this paper aims at a more homogeneous and high level classification of the verb synsets, mapping the verbs to domain independent concepts representing only temporal particulars.

Regarding semantic annotation, many tagging approaches have been proposed in the last years, ranging from lexical to semantic annotation features. Lexical annotation, such as the ones based on POS (part-of-speech) tags and syntactic roles \cite{manning2014stanford} are largely employed and serve as a basis for more complex annotation tasks. Besides assigning a tag for each single word in a sentence, lexical annotation can also cover the relationships between words, like syntactic dependency and co-reference \cite{manning2014stanford}. On the other hand, semantic annotation focuses on capturing the meaning of words and the kind of information they carry. Among the most common semantic annotation techniques are the Named Entity Recognition \cite{nadeau2007survey}, focused on recognizing numeric expressions and entities identified by proper nouns, the Sentiment Annotation \cite{manning2014stanford}, which classifies words as positive, negative or neutral for Sentiment Analysis tasks, and Semantic Role Labeling \cite{martin2000speech}, intended to determine the role of entities which refers to a given event.

Foundational ontology-based tagging is a semantic annotation task, as it aims at identifying the most primary meaning of a concept, that is, what its most basic category is. The semantic annotator most closely related to FO tagging is the SuperSense Tagger \cite{ciaramita2006broad}. SST treats the problem of super sense tagging as a sequential labeling task and implements it as a Hidden Markov Model. The tagset is composed of 41 WordNet high-level noun and verb synsets, called \textit{super senses}. It is also intended to determine the concept's primary category, but the set of super senses can be considered inconsistent, as it mixes higher- and lower-level concepts all together, with overlapping categories that would allow multiple possibilities of classification. For example, a ``cake'' could be classified as ``foods and drinks'', but also as ``man-made objects'', as these super senses overlap, being the first a specialization of the second. FO tagging, in turn, adopts a more rigorous and formal semantic meta-model, pushing the classification to a more stable and conceptually grounded set of categories.	

\section{Verbs and Ontologies}\label{verbonto}
At first sight, it may seem unsuitable to fit verbs in a classification driven by categories from the DOLCE ontology. An attempt to apply the OntoClean methodology \cite{guarino2009overview} metaproperties and constraints will prove challenging, making it hard to assign rigidity, unity and identity values to verbs, since they are not entities by themselves. On the other hand, the occurrence of a verb in fact represents a temporal entity, in general given by a noun, for which it is possible to assign the aforementioned metaproperties and restrictions. For example, the occurrence of the verb \textit{run} leads to a \textit{running}, \textit{appear} to \textit{apparition}, \textit{leak} to \textit{leakage}, and so on. The proposed approach is to track back the noun denoting a temporal entity that best represents a verb's occurrence, and which is already mapped to DOLCE, and map the verb to the same DOLCE class that was assigned to its noun counterpart.

The adopted model for introducing verbs in the DOLCE-oriented WordNet ontology finds support in the ITP (Intelligent Text Processing) linguistic ontology proposed by Dahlgren \cite{dahlgren1995linguistic}, which is a content ontology for natural language processing that intends to represent a ``world view'' based on assumptions about what exists in the world, including verbs, viewed as essential elements in NLP, and how to classify it. Regarding verbs, the ITP ontology follows the Vendlerian approach \cite{vendler1967linguistics}, classifying each verb as an event or state, being events further subdivided into activities, accomplishments and achievements. This is very close to the classification of temporal entities defined in DOLCE, which presents some small variations, detailed in the next Section.

\subsection{Perdurants in DOLCE}\label{dolceperds}
The most fundamental distinction in DOLCE is that between \textit{endurants} and \textit{perdurants}. Simply put, endurants are entities that are fully present (that is, all their parts are present) at any moment they are present. Contrarily, perdurants are entities that span in time, being only partially present at a given moment, as some of their parts (past or future phases) may not be present at that time. The classification of verbs is restricted to the perdurant branch, which is subdivided into the \textit{stative} and \textit{event} classes, informally described next. A formal description of the DOLCE concepts can be found in \cite{masolo2003wonderweb}.

A temporal entity, i.e., a perdurant, is considered \textit{stative} if it is cumulative, and \textit{eventive} otherwise. For example, a ``sitting'' is stative because it is cumulative, since the sum of two sittings is still a sitting. Stative occurrences are subdivided into \textit{states} and \textit{processes}. A state is an occurrence whose all temporal parts can be described by the same expression used to describe the whole occurrence. A ``sitting'' is a state because all of its temporal parts are also sittings. Differently, ``smoking'' is a process, because some temporal parts of a smoking are not smokings themselves. A state can be further specialized into a \textit{cognitive-state}, that is, a state of the (embodied) mind.

Events are subdivided into \textit{accomplishments}, \textit{achievements} and \textit{cognitive-events}. An occurrence is called an achievement if it is atomic, and an accomplishment otherwise. A cognitive-event is an event occurring in the (embodied) mind. Accomplishments are also further subdivided into a series of more specific concepts, but for the purposes of this work only the higher level classes are of interest.

\subsection{DOLCE Lite Plus vs. DOLCE Ultra Lite}\label{ontocomp}
The alignment between WordNet noun synsets and DOLCE was recently updated to address an entity typing task \cite{gangemi2012automatic}. Called OntoWordNet (OWN) 2012, this update builds upon OWN \cite{gangemi2003ontowordnet}, an OWL version of WordNet-DOLCE alignment. Besides revising the manual mappings, OWN 2012 also adopts a different version of DOLCE, the DOLCE Ultra Lite Plus \cite{gangemi2008norms}, which is a simplified version of DOLCE Lite Plus (called simply DOLCE in the rest of this work), intended to make classes and properties names more intuitive and express axiomatizations in a simpler way, among other features. An additional lightweight foundational ontology, called DOLCE Zero (D0), was also developed and integrated into DULplus, generalizing some of its classes.

A substantial difference that can be noted between DLP and the DULplus+D0 resulting ontology (herein called simply DULplus) refers to their hierarchical organization. As mentioned in Sec. \ref{dolceperds}, the main branches in DLP are \textit{endurant} and \textit{perdurant}, and this is the most fundamental distinction that guided the ontology development. DULplus adopts a more relaxed hierarchy, where the distinction between endurants and perdurants is left aside. Instead, there is a highest level class called \textit{Entity}, whose direct subclasses are \textit{Abstract}, \textit{Cognitive Entity}, \textit{Event}, \textit{Information Entity}, \textit{Object}, \textit{Quality}, \textit{System} and \textit{Topic}. The \textit{Event} class is subdivided into \textit{Action} and \textit{Process}, and there is no equivalent to the DLP class \textit{State}. Also, it's not clear whether \textit{Cognitive Entity}, defined as ``Attitudes, cognitive abilities, ideologies, psychological phenomena, mind, etc.'' refers to endurants, perdurants or both.

Clearly identifying which noun synsets in WordNet denote temporal entities is a key point in the methodology adopted in our verb classification, since we consider only perdurants as suitable categories for verbs. Given the above mentioned characteristics of DULplus, even though OWN 2012 is a more recent resource, we opted for using the original WordNet-DOLCE alignment, because we believe that the more rigorous conceptualization expressed in the DLP hierarchy could provide us with a higher quality verb classification.

\section{Alignment Methodology}\label{method}
To carry out the WordNet verb synsets alignment to the DOLCE concepts, the noun synsets alignment provided in \cite{gangemi2003sweetening} was used as a reference frame from which the DOLCE classes were transferred to the related verbs, according to the relevant links between word senses in WordNet. The alignment methodology comprises the following steps:

\noindent \textbf{Update and Expansion of Nouns Alignment:} as available in \cite{masolo2003wonderweb}, 813 noun synsets have been aligned to 50 DOLCE classes. We updated these 813 alignment to bring them to a more recent version of WordNet, since the original ones were done over version 1.6. Using the synset ID mappings provided by Daud\'{e} et al. \cite{daude2003making}, the alignments were migrated from version 1.6 to version 3.0, resulting in 809 aligned synsets (some synsets are excluded or merged from one version to the subsequent one). Then, we expanded the alignments to assign the DOLCE classes to the remaining synsets. The 809 aligned synset are located at the highest levels of the WN hierarchy, then, using the hypernym and instance links recursively, the synsets at the lower levels inherited the DOLCE class from their parent synsets. The final alignment contains 80,897 noun synsets, which corresponds to 98.5\% of the WN 3.0 noun database. The remaining 1.5\% includes the synsets that were not considered in the original DOLCE-WN alignment, and their hyponyms and instances.

\noindent \textbf{Top Level Verbs Selection:} similarly to the nouns alignment, the verbs classification was performed over the top level synsets, to be later propagated down the taxonomy through the hyponym links. The WordNet verb taxonomy contains 560 top level synsets, that is, synsets that have no hypernyms, and that were selected as candidates for the direct alignment.

\noindent \textbf{Direct Links:} for each of the 560 top level verb synsets, we retrieved all the related word senses given by the \textit{derivationally related form} lexical link. The derivationally related words were then manually filtered in order to identify, among them, the noun that would represent the verb occurrence. In general, these are words whose definition (gloss) starts with expression such as ``the act of'', ``the process of'' or ``the state of'', which are strong evidences that they are classified as perdurants. For example, for the verb \textit{move} (``be in a state of action''), the derivationally related words retrieved were \textit{motion}, \textit{move} and \textit{mover}. In the manual analysis phase, the word \textit{move} (``the act of deciding to do something'') was identified as the correct noun counterpart for the verb move, and the DOLCE class \textit{event} associated to it was then also assigned to the verb.

\noindent \textbf{Indirect Links:}  since not all verbs have a suitable noun counterpart, in many cases no direct link can be found. For the verbs that have no derivationally related form, or none of the derivationally related forms represents the verb occurrence, an indirect path to the appropriate noun was manually searched. This path relies on other kind of links, such as \textit{antonym} and \textit{verb group}. The verb group link acts in a way similar to the derivationally related form, but linking only verbs among them, and the antonym link is also useful because, in general, to be comparable, the verbs need to be of the same kind, leading to the same DOLCE category. This is not always true, as some states, for example \textit{stand still}, have as antonym an event, in this case, \textit{move}, so an additional manual check is required to ensure that the found path is indeed valid. As an example, the verb \textit{ignore} (``be ignorant of or in the dark about''), which has no suitable derivationally related forms, has as antonym the verb \textit{know} (``be cognizant or aware of a fact or a specific piece of information''), which is, in turn, linked to the noun \textit{knowingness} (``having knowledge of''), classified as a \textit{cognitive-event}, so that DOLCE class was directly assigned to the verb \textit{know} and, as a consequence, indirectly linked to the verb \textit{ignore}.

\noindent \textbf{Manual Assignment:} finally, for the verbs for which no explicit direct or indirect link to a noun could be found, a careful manual evaluation was carried out. This evaluation has taken into account the implicit relationship to other (preferably already classified) verbs, given by the words present in each synset's gloss. Using the gloss to uncover implicit relationships is an important procedure to make the classification as less subjective as possible. For example, verbs having glosses beginning with ``be'', like \textit{wear} (``be dressed in''), \textit{stay in place} (``be stationary'') or \textit{sit} (``be seated'') are strong \textit{state} candidates. In other cases, even if the link is not explicit in WordNet, the relationship is very clear, for instance, between the verbs \textit{arch} (``form an arch or curve'') and \textit{overarch} (``form an arch over''). Since \textit{arch} was, by inheritance, mapped to the \textit{event} class, and given the high similarity between their glosses (indeed, the second could possibly be a specialization of the first), we could also classify \textit{overarch} as an \textit{event}. In all scenarios, checking if the concept's characteristics described in Sec. \ref{dolceperds} apply and further analyzing the verb's hyponyms to make sure they also fit in the chosen DOLCE category helped us to reach a consistent classification.

\begin{figure*}[t]
	\centering
	\includegraphics[width=4.8in, height=0.9in]{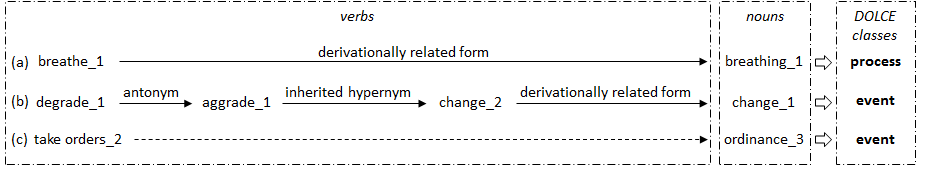}
	\caption{Examples of WordNet-DOLCE verb mappings obtained by (a) direct link, (b) indirect links and (c) manual assignment. Full lines stand for explicit links in WordNet, while the dashed line represents an implicit relationship. The numbers indicate the word sense in WordNet 3.0}
	\label{fig:mapping_examples}
\end{figure*}

Figure \ref{fig:mapping_examples} shows some examples of mappings between WordNet verbs and DOLCE concepts, reached through explicit direct and indirect links, and following implicit relationships identified by manual evaluation.

\section{Alignment Results}\label{alignres}
After applying the alignment methodology described in Sec. \ref{method}, the 560 top level verb synsets were mapped to five DOLCE classes: \textit{event}, \textit{state}, \textit{process}, \textit{cognitive-event} and \textit{cognitive-state}. These are the same classes used previously to classify all the WordNet noun synsets denoting perdurants. A total of 52.5\% of the synsets were mapped through explicit links, being 36.25\% direct links and 16.25\% indirect links to noun synsets, and 47.5\% through implicit relationships identified by manual analysis. The top level mappings were then propagated down the verb taxonomy using the hypernym-hyponym links, and all the troponyms, as verbs' specializations are called in WordNet, inherited their parent's DOLCE class, resulting in a 100\% mapped verb database.

The adequacy of WordNet hypernym links to effectively represent subsumption relationships is a common concern, but, since we are dealing with very high level categories, the probability of errors in the taxonomy propagation, although not completely eliminated, is considerably reduced. Although not 100\% of the WN hypernym links can be considered correctly assigned, they are intended to represent subordination \cite{miller1995wordnet}, meaning that synsets are linked in the hierarchy because they somehow show a strong similarity regarding their nature, that is, what they primarily represent in the world, and then tend to converge to the same upper class even if they don't follow a strict subsumption relationship.

Table \ref{tab:result} shows the final distribution of both top level and full taxonomy synsets over the five DOLCE classes. Although it may seem unbalanced towards the \textit{event} class, this is coherent with the noun synsets mapping, where 75\% of the 8,522 perdurant nouns were also classified as \textit{events}. A possible reason for these figures is that most verbs describe actions, and \textit{action}, described as ``a perdurant that exemplifies the intentionality of an agent'' \cite{masolo2003wonderweb}, is indeed a subclass of \textit{accomplishment}, which is a subclass of \textit{event}. The original WordNet-DOLCE nouns alignment opted for a higher level mapping, keeping at the \textit{event} class instead of drilling down more specific concepts, and this choice is also reflected in our final verb mapping.

\begin{table}
	\centering
	\caption{WordNet-DOLCE verb synsets alignment statistics}
	\begin{tabular}{lcc}
		\textbf{DOLCE class\,\,} & \textbf{\,\,Top Synsets\,\,} & \textbf{\,\,Full Taxonomy} \\ 
		\hline\noalign{\smallskip}
		event & 412 & 12,037 \\ 
		cognitive-event & 63 & 854 \\ 
		state & 62 & 597 \\ 
		process & 15 & 259 \\ 
		cognitive-state & 8 & 20 \\ 
		\hline\noalign{\smallskip}
		\textbf{Total} & \textbf{560} & \textbf{13,767}
	\end{tabular}
	\label{tab:result}
\end{table}

\section{Evaluation}\label{eval}
To evaluate the usefulness of the resulting alignments in a semantic annotation task, we run experiments using two datasets: the SemCor dataset \cite{miller1993semantic} and the eXtended WordNet \cite{moldovan2001logic}. SemCor is a subset of the Brown corpus \cite{ku1967computational} which has been manually annotated with WordNet sense numbers. SemCor 3.0 is annotated with WordNet 3.0 synsets, and is divided in three parts: ``brown1'' and ``brown 2'', where all nouns, verbs, adjectives and adverbs are annotated, and ``brownv'', with annotations only for verbs. We used the ``brown1'' and ``brown2'' subsets, which make up a total of 20,132 sentences (the original dataset contains 20,138 sentences, but 6 of them are empty sentences). The eXtended WordNet, or XWN, is a resource that provides logical forms for all WordNet synset glosses. Besides the logical forms, XWN also includes word sense disambiguation, being all words present in every gloss annotated with its correspondent WordNet sense number. XWN is divided into four files, one for each grammatical class: ``noun'', ``verb'', ``adj'' and ``adv''. We used the ``noun'' and ``verb'' datasets, since these are the synsets covered in our alignment. To build the sentences, the synset head word, that is, the first word in the synset, was followed by the word ``is'' and concatenated with the synset gloss, making up a total of 93,197 sentences.

Using the sense number information available in both datasets, we retrieved the synset ID for each word using the JWI API \cite{finlayson2014java}, and identified the DOLCE class associated with that ID. All the nouns and verbs in a sentence were labeled with its DOLCE class, and all the adjective, adverbs and stop words received a null label. The full labeled datasets were then used as a gold standard in the semantic annotation task evaluation.

To perform the FO tagging, we opted for the first sense heuristic, or Most Common Sense, as word sense disambiguation technique, using the WordNet's sense ranking to retrieve the most frequent sense for each word. Although being a very straightforward technique, MCS outperforms many WSD systems \cite{mccarthy2004finding} and is a good alternative for disambiguating commonsense data.

Our semantic annotator then received as input all the SemCor and XWN words/phrases, grouped into sentences, and identified the synset ID using the MCS WSD heuristic, labeling each token with a DOLCE class or with the null label when the synset ID wasn't found in the synset-class mapping. The results were contrasted with the gold standard and compared with two baselines: the random baseline, and the SuperSense Tagger. Although the comparison with SST is only possible for the SemCor dataset, we believe it is worth to show the difference in the results, since, as mentioned in Sec. \ref{relwork}, this is the semantic annotation approach that is most similar to our foundational ontology-based tagging.

The results\footnote{Computed by the ``conlleval'' script, available at http://goo.gl/YL2IBz} are summarized in Tab. \ref{tab:eval}. The first line shows the accuracy of a baseline that, for each word/phrase, chooses a sense number at random and then assigns the correspondent DOLCE label. The efficacy of the MCS disambiguation method adopted by the FO tagging can be observed by the F1-Score for both datasets, well above the random baseline. When compared to the SST, FO Tagging presents an increase of 9.05\% in the F1-Score for the SemCor dataset.

\begin{table}
	\centering
	\caption{Evaluation results}
	\begin{tabular}{lccc|ccc}
		~                  & \multicolumn{3}{c}{\textbf{XWN}} & \multicolumn{3}{c}{\textbf{SemCor}} \\
		~                  & \textit{Precision} & \textit{Recall} & \textit{F1-Score} & \textit{Precision} & \textit{Recall} & \textit{F1-Score} \\ \hline
		Random             & 71.82     & 72.04  & 71.93    & 61.52     & 62.52  & 62.02    \\
		FO Tagging         & \textbf{89.68}     & \textbf{89.74}  & \textbf{89.71}    & \textbf{86.10}     & \textbf{86.36}  & \textbf{86.23}    \\
		SuperSense Tagging & -         & -      & -        & 76.65     & 77.71  & 77.18    \\
	\end{tabular}
	\label{tab:eval}
\end{table}

It is important to emphasize that the goal of the evaluation is not to judge the quality of the alignment itself, but rather to assess how it could be effectively used in an annotation task. Considering the final mapping as a standard to be followed (which is also reflected in the gold standard, built based on it), the bottleneck stands in finding the correct label for a word/phrase when it has more than one label associated to it. The random baseline is relatively high because it does not mean choosing a label among all existing ones, but randomly picking a label only among the ones associated with a given word/phrase. The results, then, show the accuracy of the chosen approach for FO tagging at selecting the most suitable label from the standard mappings set.

\section{Conclusion}\label{concl}
The previous effort to align WordNet to the foundational ontology DOLCE led to a conceptually more rigorous version of the WordNet noun taxonomy, meant to increase its adequacy as an ontology. We presented an extension to this alignment, using it as a reference frame to map also the verb synsets, using explicit links between word senses and implicit relationships given by the words present in the verbs' glosses to track back the noun that would best represent an occurrence of a given verb, and assign to this verb the same DOLCE class previously associated to its noun counterpart. After aligning the 560 top level verb synsets, the classification was expanded through the taxonomy, resulting in a 100\% aligned WordNet 3.0 verb database.

The resulting alignment was then used in the implementation of a semantic annotation framework, the FO Tagging, which used the Most Common Sense word sense disambiguation technique to identify to which synset each word belongs to and subsequently retrieve the DOLCE label associated with that synset. Compared to the SuperSense Tagger, the most similar semantic annotation tool, FO Tagging shows an increase of 9.05\% in the F1-Score for the SemCor dataset. In addition to the increase in the accuracy, FO Tagging also introduces an ontologically well-grounded set of categories, pushing the classification to a higher level than that provided by the SST tagset.

Besides contributing to expand the benefits brought by the initial noun mapping, the introduction of the verb alignment can also help in the execution of semantic tasks involving natural language processing, like text entailment and question answering, where concepts need to be mapped to a smaller set of categories in order to reduce the reasoning search space. DOLCE classes provide a suitable semantic representation for such tasks, and the evaluation has shown that, even with a straightforward word sense disambiguation technique, with the aid of the WN-DOLCE alignment it is possible to annotate text with a high accuracy. As future work, we intend to try more sophisticated WSD methods to improve the robustness of our semantic annotator, as well as start the analysis of the adjective and adverb databases to expand the alignment also to those synsets. Furthermore, this foundational ontology-based annotation tool will be integrated into a text entailment mechanism currently under development. This mechanism links the text \textit{T} to the hypothesis \textit{H} using dictionary definitions as intermediates, and try to determine whether H can be obtained from T by means of a sequence of transformation operations, performed over their foundational representations obtained through FO tagging.

\begin{flushleft}
\begin{tabular}{p{0.75\linewidth} p{0.2\linewidth}}
	\vspace{0pt}
	\textbf{Acknowledgments.} This work is in part funded by the SSIX Horizon 2020 project (grant agreement No 645425).
	& 
	\vspace{0pt}
	\raggedleft \includegraphics[width=0.6in, height=0.4in]{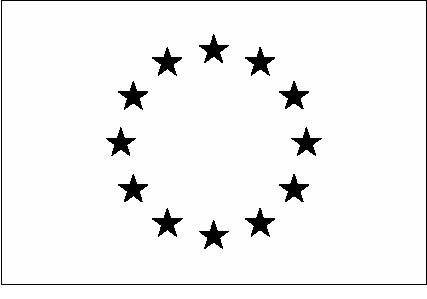}
\end{tabular}
\end{flushleft}

\bibliography{ekaw2016}
\bibliographystyle{splncs03}

\end{document}